%% file: main.tex
\documentclass{article}
\usepackage[dblblindworkshop, final]{neurips_2026}

\usepackage[utf8]{inputenc}
\usepackage[T1]{fontenc}
\usepackage[hyphens]{url}
\usepackage{booktabs}
\usepackage{array}
\usepackage{amsfonts}
\usepackage{amsmath}
\usepackage{amsthm}
\newtheorem{proposition}{Proposition}
\newtheorem*{mainprop}{Proposition~1}
\usepackage{nicefrac}
\usepackage{microtype}
\usepackage{xcolor}
\usepackage{graphicx}
\usepackage{algorithm}
\usepackage{algpseudocode}
\usepackage[capitalize]{cleveref}
\crefformat{equation}{Eq.~#2#1#3}
\crefname{appendix}{Appendix}{Appendices}
\graphicspath{{figures/}}
\definecolor{revcolor}{rgb}{0,0,0}
\newcommand{\rev}[1]{{\color{revcolor}#1}}
\definecolor{revbcolor}{rgb}{0,0,0}

\definecolor{revccolor}{rgb}{0,0,0}
\newcommand{\revc}[1]{{\color{revccolor}#1}}

\title{\PCRCTitle}
\author{%
  Siddharth Aphale \\
  Stanford University \\
  \texttt{saphale@stanford.edu} \\
  \And
  Ayushman Singh\thanks{Work done independently outside Sesame.} \\
  Sesame AI \\
  \texttt{ayushman@sesame.com} \\
}

\workshoptitle{8th Robot Learning Workshop: Is Physical AI Going Zero-Shot?}
\makeatletter
\renewcommand{\@trackname}{\@neuripsordinal\ Conference on Neural Information Processing
  Systems (NeurIPS \@neuripsyear). Workshop: \@workshoptitle}
\makeatother

\input{generated/names}

\begin{document}

\maketitle

\begin{abstract}
\revc{Sparse-reward reinforcement learning often fails because rollouts from the unassisted
evaluation start rarely reach later task stages. Reverse curricula address this by starting some
training rollouts from easier intermediate states, called scaffolds. The main choices are which intermediate starts to use (\emph{scaffold access}) and how to distribute training across them (\emph{scaffold allocation}). Most prior
curricula pace removal on one shared schedule, which can fail when task instances, or
contexts, learn at different rates. We introduce an online, learner-agnostic
reset controller that gives every context its own curriculum. We use binary rollout success to decide whether to move a context's start forward or back. A stalled frontier also gets an occasional harder-start trial. A counting construction shows that
synchronized global pacing can be insufficient
when contexts need conflicting amounts of assisted practice. Across six navigation and
manipulation settings, scaffold access improves learning and enables success in three where
unassisted training fails within the reported budget. None of the global schedules we tested solved both groups in the constructed conflict. The per-context controller did. Group-level pacing works if the labels capture how contexts learn. Without those labels, per-context pacing still handles differences within a group. These results suggest pacing assistance separately for contexts that learn at different rates.}
\end{abstract}

\section{Introduction}
\label{sec:intro}
An agent cannot learn later task stages if it rarely reaches them. In sparse-reward
manipulation, a policy that cannot grasp an object receives no signal about placing it on a
shelf. Hindsight relabeling reuses visited states, but it cannot provide experience in stages
the policy never reaches~\citep{andrychowicz2017her}. Simulators with state restoration offer
another option: start some training rollouts from useful intermediate states.

We call a temporary, easier start a \emph{scaffold}. Several ordered scaffolds form a ladder
from a state near success to the unassisted evaluation start. We call the unassisted evaluation
start the \emph{target start}. A reverse curriculum chooses a ladder level for each training
rollout and gradually removes assistance~\citep{florensa2017reverse,florensa2018goalgan}.
This creates two decisions. We separate the choice of available intermediate starts (\emph{scaffold access}) from how training is distributed across them (\emph{scaffold allocation}). We call each task instance, such as an object--goal pair, a \emph{context}.

Prior reverse curricula have mainly studied scaffold access. ACED and RFCL also progress
separately along each demonstration~\citep{dai2021aced,tao2024rfcl}. However, they do not
study which contexts should share a curriculum pace. One global pace can be too coarse when
contexts learn at different rates. The on-table contexts are ready for target-start practice early, but the in-air contexts still need rollouts that begin with the object already grasped.
The hard cases stall if scaffolds disappear too soon. Keep them too long, and the easy cases waste rollouts. Average success can conceal this failure. A counting construction shows that some
two-context problems cannot be solved by any synchronized global schedule
(\cref{prop:sync}).

We introduce an online reset controller with one frontier per context.
The controller uses recent success to move a context toward the target start and failure to move it back. If progress stalls, it tries one harder level. The controller observes only
binary rollout success. The reward, optimizer, and learner remain unchanged. The controller also needs no group labels.

Our experiments separate scaffold access from scaffold allocation. Across six navigation and
manipulation settings, scaffold access improves learning. It enables success in three settings
where target-start training fails within the reported budget. All the global schedules we tested failed on at least one group in the constructed conflict. Our controller solved both. Known groups are useful when their members need roughly the same pacing. Otherwise, a shared frontier can fail. Our controller remains strong in both cases. Taken together, the results suggest pacing assistance separately for contexts that progress at different rates.

\begin{figure}[t]
  \centering
  \includegraphics[width=\linewidth]{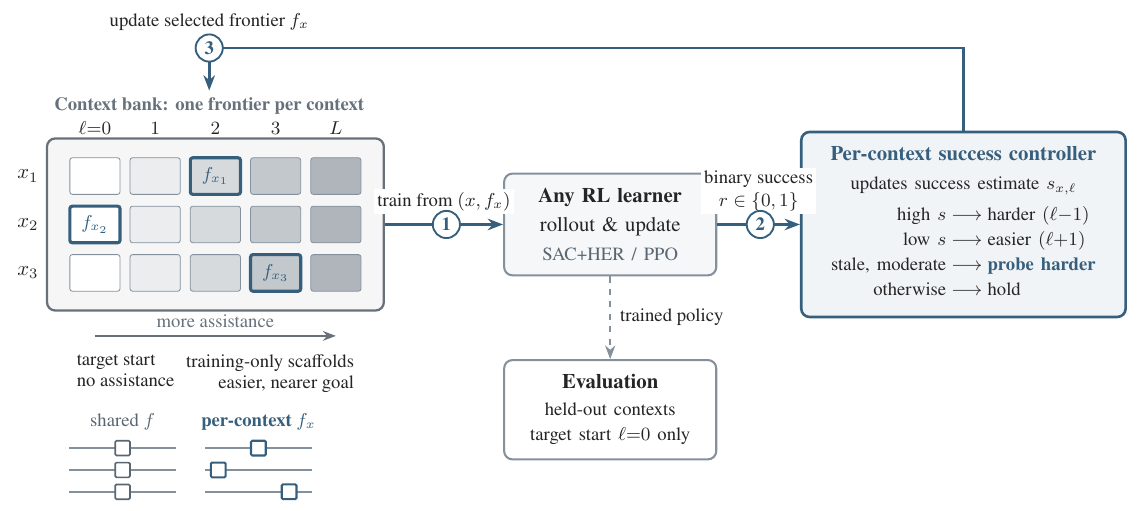}
  \caption{\textbf{Our controller maintains per-context scaffold frontiers.} Rows are contexts;
  columns are training-only levels. Level $\ell{=}0$ is the held-out evaluation start. Binary
  success triggers advance, retreat, or a cautious harder test.}
  \label{fig:method}
\end{figure}

\section{Related Work}
\label{sec:related}
\paragraph{Reverse and start-state curricula.} Reverse Curriculum Generation expands starts
outward from the goal, while GoalGAN samples goals near the learner's frontier
\citep{florensa2017reverse,florensa2018goalgan}. Other methods select starts using performance
estimates, demonstrations, or reset diversity
\citep{wohlke2020startstate,duan2025cago,yin2026diverseresets}. These methods mainly improve
scaffold access. ACED and RFCL instead advance a frontier along each demonstration
\citep{dai2021aced,tao2024rfcl}. They are the closest prior mechanisms to our controller. Our question is
which contexts should share a curriculum pace. We compare global, group-level, and per-context
pacing. Our controller indexes its frontier by task context, and its ladder need not follow trajectory
time. It also permits retreat after failure, which matters on coarse ladders.

\paragraph{Task curricula.} Teacher--student methods select tasks by learning progress,
difficulty, or regret
\citep{graves2017automated,matiisen2020tscl,portelas2020alpgmm,narvekar2020curriculum}.
Self-paced RL moves the training distribution toward the target \citep{klink2020selfpaced}.
PLR, PAIRED, and ACCEL select or generate levels near the frontier of learnability
\citep{jiang2021plr,dennis2020paired,parkerholder2022accel}. DISCOVER and ACDC build curricula
for sparse-reward reaching and manipulation \citep{discover2025,acdc2026}. Prior methods choose which tasks to train on. Here we keep the context bank fixed and choose how much reset assistance each context gets.

\paragraph{Goal-conditioned learning.} Our continuous experiments use SAC with Hindsight
Experience Replay \citep{haarnoja2018sac,andrychowicz2017her}. Relabeling makes use of states the policy has already visited, but later task stages remain out of reach if the policy never gets there. Scaffolds give the learner experience in those stages.

\section{Method}
\label{sec:method}
\paragraph{Problem setup.} Let $\mathcal{X}$ denote a finite training context bank, and let
$\ell \in \{0, \ldots, L\}$ index scaffold levels. A reset operator $G(x,\ell)$ maps context $x$
and level $\ell$ to an initial-state distribution $\rho^{x,\ell}_0$. Level $\ell=0$ is the
unassisted target distribution used for evaluation. Larger levels are training-only starts.

We call each pair $(x,\ell)$ a context-scaffold cell. Our controller maintains one active frontier $f_x$
per context, initialized to the easiest level $L$. Training draws resets from active cells
$(x,f_x)$. Evaluation uses disjoint held-out contexts at $\ell=0$ (\cref{fig:method}).

\paragraph{Controller statistics.} Our controller observes binary rollout success and selects the next
training cell. Because the policy changes during training, our controller uses recent evidence. For each
active cell, it maintains the exponential moving average
$s_{x,\ell}\leftarrow(1-\beta)s_{x,\ell}+\beta r$, where $r\in\{0,1\}$. It also tracks an
evidence count, a transition cooldown, a recency counter $c_{x,\ell}$, and a staleness counter
$d_x$. The last quantity counts eligible updates since the frontier moved. A frontier decision
requires enough evidence and a completed cooldown.

\paragraph{Active cell sampling.} Each controller step samples active cells using
\begin{equation}
\mathrm{score}(x) \;=\; \lambda_h\, s_{x,f_x}\,(1-s_{x,f_x})
\;+\; \lambda_p\,\frac{L-f_x}{L}
\;+\; \lambda_{\mathrm{stale}}\,\log\!\big(1+c_{x,f_x}\big),
\label{eq:score}
\end{equation}
The score favors cells with moderate success and gives an additional boost to less assisted or rarely sampled cells. A softmax over these scores has a uniform
exploration floor $\epsilon$ and per-cell cap $q_{\max}$. We fix all coefficients across settings
and conditions.

\paragraph{Frontier transition.} After the evidence and cooldown conditions hold, each context
updates its frontier by
\begin{equation}
f_x \leftarrow
\begin{cases}
\max(0,f_x-1), & s_{x,f_x}\ge \tau_{\mathrm{high}},\\
\min(L,f_x+1), & s_{x,f_x}\le \tau_{\mathrm{low}},\\
\max(0,f_x-1), & d_x\ge N_{\mathrm{stale}}\ \land\ s_{x,f_x}>\tau_{\mathrm{low}},\\
f_x, & \text{otherwise}.
\end{cases}
\label{eq:transition}
\end{equation}
Success pushes the frontier toward the target start. Failure sends it back. If neither happens for long enough, we try a harder level. Without this \textbf{antistall} transition, a context between the two thresholds can remain there
indefinitely. After $N_{\mathrm{stale}}$ eligible updates, our controller tries one harder start. If that
step is premature, the failure transition restores assistance. \cref{alg:pcrc} gives the
controller loop.

\begin{algorithm}[t]
\caption{Per-context curriculum controller}
\label{alg:pcrc}
\begin{algorithmic}[1]
\State \textbf{Input:} context bank $\mathcal{X}$, levels $\{0,\dots,L\}$, reset operator $G$
\Statex \hspace{\algorithmicindent}\phantom{\textbf{Input:}} thresholds
        $\tau_{\mathrm{low}},\tau_{\mathrm{high}}$, stale limit $N_{\mathrm{stale}}$
\State Initialize $f_x \gets L$, success averages, evidence counts, cooldowns, stale counters
\For{each training iteration}
  \State sample context $x$ from the active cell distribution (\cref{eq:score}) with an exploration floor
  \State reset environment from $G(x,f_x)$; collect rollout; update the learner (SAC+HER / PPO)
  \State update success average and evidence for $(x,f_x)$
  \For{each context whose evidence and cooldown conditions hold}
    \State apply the frontier transition of \cref{eq:transition}
  \EndFor
\EndFor
\State \textbf{Evaluate only on held-out contexts at level $\ell=0$.}
\end{algorithmic}
\end{algorithm}

\begin{figure}[t]
  \centering
  \includegraphics[width=\linewidth]{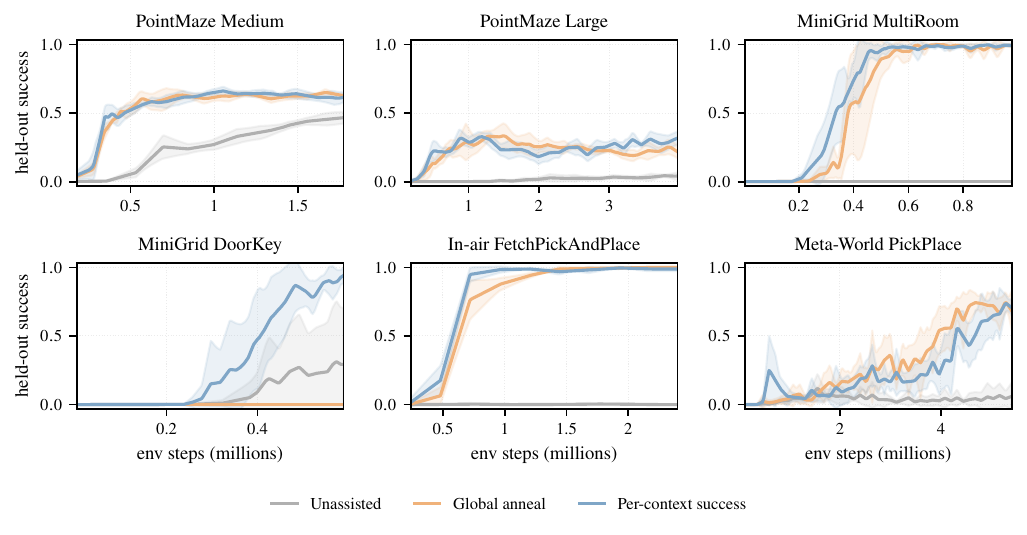}
  \caption{\textbf{Held-out target-start success across six natural settings.} Curves show mean
  $\pm$ s.d.; step ranges vary by panel. Gray is target-start training, orange is global
  annealing, and blue is \PCRCPerContextSuccess{}.}
  \label{fig:sixtask}
\end{figure}

\paragraph{Trajectory-derived scaffold construction.} We construct $G$ automatically from one
successful reference trajectory $\tau_x=(s_0,\ldots,s_T)$ per training context. We define a
progress coordinate $p(t)\in[0,1]$. We serialize the simulator state nearest each of five fixed
quantiles: $\{0,0.25,0.5,0.75,0.9\}$. For discrete DoorKey, the final quantile is $1.0$, placing
the easiest start next to the goal. Level $\ell=0$ is the target start, and larger $\ell$ lie
later on $\tau_x$. This construction uses no semantic stage labels and retains the original
sparse reward. We validate each state before RL begins. The retained gain ratio
$R_{\mathrm{retain}}$ is the fraction of the hand-built scaffold gain recovered by this
construction. A value $\geq 1$ matches or exceeds the hand-built scaffold. The procedure
requires reference trajectories and reset access. It does not discover scaffolds from scratch.

\paragraph{\revc{A structural limit of synchronized pacing.}} A counting construction shows that
global schedules can fail when contexts need different amounts of assisted practice.

\begin{proposition}[Synchronized global pacing can be insufficient]
\label{prop:sync}
Every $K>1$ admits a two-context, two-level scaffold curriculum with per-context budget $K+1$
such that a context-local schedule solves both contexts, while every synchronized global schedule fails at
least one context.
\end{proposition}

A synchronized schedule gives both contexts the same number of assisted episodes. In the
construction, one context needs one assisted episode and the other needs $K$ (full proof in the
supplement). The construction rules out schedules that keep contexts synchronized. A global mixture can still retain several scaffold levels. We therefore test an asynchronous global mixture. This is a counterexample for synchronized pacing. It says nothing about whether our controller is optimal.

\section{Experimental Setup}
\label{sec:setup}
We evaluate our controller on six natural settings and three pacing conflicts. The conflicts are a
constructed Fetch PickAndPlace case, an unmodified Meta-World pair, and a within-group case
(Results). We keep the controller fixed but match the learner to each environment. We use
SAC+HER for continuous navigation, Fetch, and Meta-World PickPlace~\citep{yu2020metaworld}.
MiniGrid uses PPO with a partial-view convolutional encoder~\citep{chevalier2023minigrid}.
Scaffolds use geodesic or route-based resets for navigation, stage-based DoorKey levels, and
reverse-curriculum starts for manipulation. Banks contain $100$ training and $50$ held-out
contexts. MiniGrid settings and the Meta-World conflict instead use $200$ and $100$.

\begin{table}[t]
  \centering
  \caption{\rev{\textbf{Scaffold access in six settings.} Median AUC, final success, and seeds
  solved ($\geq 0.5$). $\Delta$ gives ours--target difference; O denotes ours and T target-start training. DoorKey
  emphasizes final success and solved seeds; n/a means neither solved.}}
  \label{tab:sixtask}
  \small
  \resizebox{\linewidth}{!}{\input{tables/sixtask_table}}
\end{table}

The constructed conflict pairs easy on-table and hard, long-range in-air Fetch
contexts~\citep{plappert2018multigoal}. We omit middle difficulties so the groups need different
pacing. Methods share the learner, contexts, rollout workers, reward, and budget. Baselines vary
only in scaffold-level selection (\cref{tab:baselines}). Controller hyperparameters are tuned on
development seeds and fixed across settings. Evaluation uses
held-out contexts at $\ell=0$ for five or six seeds. The primary metric is held-out target-start
AUC, comparable only within a task. We also report final success, seeds solved, and worst-group
values for conflicts. Paired seed-bootstrap intervals provide the main uncertainty estimates;
sign tests are descriptive.

\paragraph{Trajectory-derived scaffold experiments.} For PointMaze Large, MiniGrid DoorKey,
and in-air FetchPickAndPlace, we replace manual ladders with the trajectory-derived construction
(Method). References use a shortest-path planner, a breadth-first solver, and a scripted
pick-and-place controller. Global annealing and random sampling use the same ladders, separating
ladder quality from pacing.

\section{Results}
\label{sec:results}

\paragraph{Scaffold access improves sparse learning across six settings.}\label{sec:res-access}
Across all six natural settings, our controller improves held-out target-start AUC over target-start
training, and every paired-seed difference is positive (\cref{tab:sixtask,fig:sixtask}). With scaffolds, the agent learns tasks that target-start training does not solve within our budget, and learns others faster. Local
allocation matters chiefly under pacing conflict.

\begin{figure}[t]
  \centering
  \includegraphics[width=\linewidth]{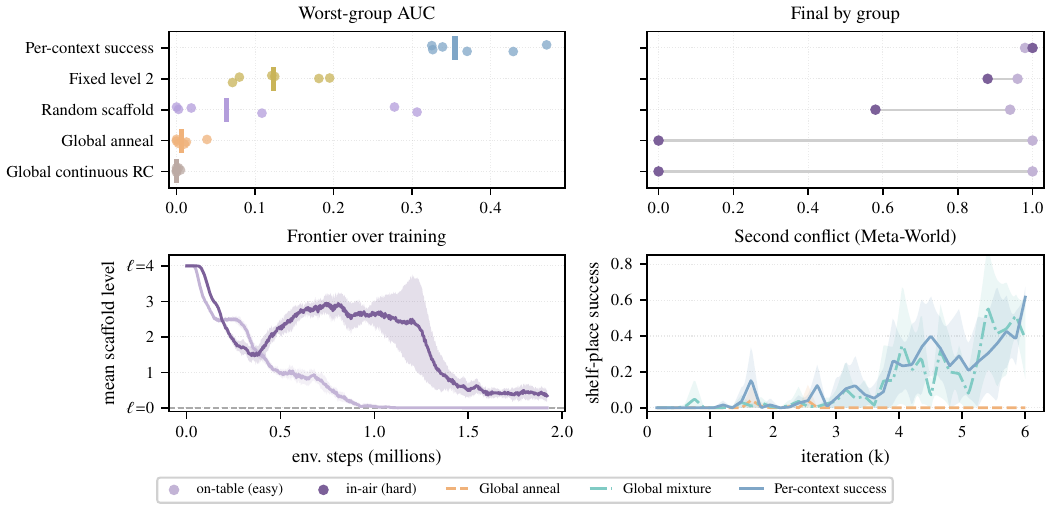}
  \caption{\textbf{Local frontiers under pacing conflicts.} Constructed-conflict panels show
  worst-group AUC, group finals, and frontier levels. Meta-World shows shelf-place success.
  Global annealing collapses while our controller persists.}
  \label{fig:conflict_main}
\end{figure}

\paragraph{Per-context allocation improves performance under pacing conflict.}\label{sec:res-conflict}
The constructed PickAndPlace conflict realizes \cref{prop:sync}. The controller reaches near-perfect success on both groups by the end of training. With global annealing or synchronized reverse curriculum, only the on-table group learns (\cref{fig:conflict_main}). Its
median worst-group AUC is $0.354$, compared with $0.123$ for the strongest fixed scaffold. A
separate six-seed replication gives
$\Delta=+0.24\;[0.18,0.38]$.
\label{sec:res-mw-conflict}
A related one-sided pacing conflict appears on unmodified Meta-World push and shelf-place
tasks. Push can still learn with assisted starts. Shelf-place needs more training from states where the object is already grasped.
Five-level global annealing ends at $0.000$ shelf-place success. Our controller reaches
$0.61\pm0.06$ and wins all six paired worst-group comparisons (\cref{fig:conflict_main}, panel 4).
Local allocation is therefore less sensitive to a coarse assistance axis.

\begin{figure}[t]
  \centering
  \includegraphics[width=0.92\linewidth]{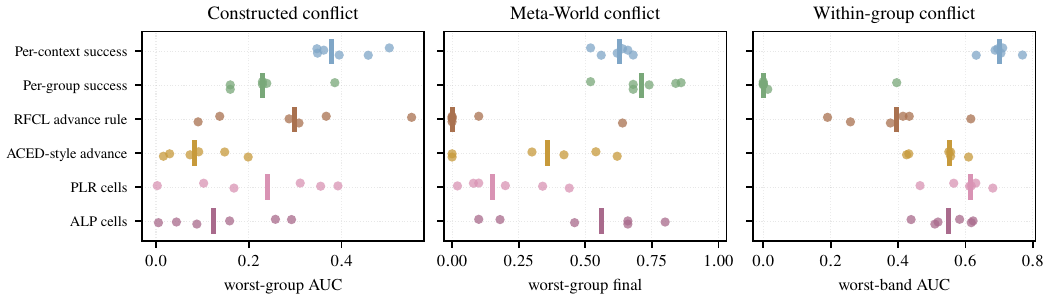}
  \caption{\rev{\textbf{Pacing granularity must match heterogeneity.} Dots are seeds; ticks are
  medians. Our controller leads the constructed conflict. Group pooling leads Meta-World. Pooling the
  single group fails on its hard band.}}
  \label{fig:granularity}
\end{figure}

\paragraph{Pacing granularity should match heterogeneity.}\label{sec:res-granularity}
On the constructed conflict, our controller is more sample-efficient than the oracle-label group frontier
(worst-group AUC $0.378$ vs.\ $0.229$; paired
$\Delta=+0.18\;[0.04,0.29]$ over six seeds). On Meta-World, the group frontier wins all six seeds
(group frontier minus our controller: $\Delta=+0.10\;[0.03,0.22]$). Pooling works when the group labels reflect actual learning differences. \label{sec:res-withingroup} It breaks down when a label lumps together contexts that learn at different rates. In a PickAndPlace bank with bimodal goal height under
one label, the pooled frontier has a worst-band AUC of $0.000$, while our controller reaches $0.701$ (paired
$\Delta=+0.69\;[0.47,0.74]$ over six seeds; \cref{fig:granularity}, right). Our controller wins every paired
worst-band AUC comparison.

\paragraph{Local, bidirectional frontiers drive the conflict gain.}
\label{sec:res-mech}
An RFCL-style ablation \citep{tao2024rfcl} keeps local frontiers but removes calibrated sampling
and antistall. The RFCL-style ablation matches our controller's worst-group AUC in both conflicts. Its final Meta-World success, though, varies more across seeds ($0.37\pm0.28$ vs.\ $0.61\pm0.06$). Local frontiers account for most of the AUC gains. Calibrated sampling and antistall make final performance more reliable. Advance-only RFCL
reaches $0.12$ on coarse Meta-World,
versus our controller's $0.61$, so retreat matters on coarse ladders.

\paragraph{Trajectory-derived scaffolds reduce manual level design.}\label{sec:res-auto}
Across three settings, trajectory-derived ladders retain or exceed hand-built gains. Their
$R_{\mathrm{retain}}$ values are $1.21\;[1.05,1.44]$, $1.47$, and
$1.06\;[1.05,1.06]$ (\cref{tab:auto}). All three automatic conditions beat
target-start training on every seed.

\section{Discussion and Conclusion}
\label{sec:discussion}
A reverse curriculum has to find useful starting states and decide how long to keep training from them.
Our controller adapts one pace per context using binary success. The controller moves a context closer to the target after sustained success. It backs off when success drops and occasionally tests a harder start when progress stalls. Its frontiers are inspectable
and require no group labels. Per-context pacing matters when learning rates vary, especially within labeled groups. Otherwise, global or pooled schedules can work well.
An exploratory contrastive-learner comparison removes the PointMaze access bottleneck but not the
Fetch bottleneck (\cref{sec:res-strong}). Scaffolds are most useful when the learner is held back by poor access to later task stages.
The method relies on restorable states that capture the task bottlenecks and can be ordered from easier to harder starts.
Contexts that learn at different rates should not be forced to share a frontier.

\paragraph{Limitations and future work.} Our controller requires restorable states, limiting use on
physical robots. Conflict prevalence and level needs for longer tasks remain open.
Future work should study automatic grouping and cross-embodiment pacing
\citep{yang2026analogies}.

\clearpage
\bibliographystyle{plainnat}
\bibliography{references}

\clearpage
\appendix
\setlength{\intextsep}{6pt plus 2pt minus 2pt}
\raggedbottom
\counterwithin{table}{section}
\counterwithin{figure}{section}
\numberwithin{equation}{section}
\crefalias{section}{appendix}
\crefalias{subsection}{appendix}

\section{Experimental protocol and implementation}
\label{app:exp-details}
Appendix~A specifies the common protocol and its setting-specific deviations;
Appendix~B describes automatic scaffold construction; Appendix~C gives the conflict proof and
supporting evidence; and Appendix~D collects controller sensitivity and boundary cases.

\subsection{Environments, scaffolds, and evaluation}

\begin{table}[H]
  \centering
  \caption{\textbf{Experimental protocol by setting.} Train/test gives the numbers of training
  and held-out contexts; seeds/eval gives primary seeds and episodes per evaluation point.
  All evaluation uses held-out target starts ($\ell=0$).}
  \label{tab:setup}
  \scriptsize
  \setlength{\tabcolsep}{3.5pt}%
  \begin{tabular}{@{}llcccc@{}}
    \toprule
    Setting & Scaffold & Learner & Train/test & Iterations & Seeds/eval \\
    \midrule
    PointMaze Medium & geodesic & SAC+HER & 100/50 & 2500 & 5/100 \\
    PointMaze Large & geodesic & SAC+HER & 100/50 & 4000 & 5/100 \\
    MiniGrid MultiRoom & route & PPO & 200/100 & 2000 & 6/50 \\
    MiniGrid DoorKey & task stages & PPO & 200/100 & 2000 & 6/50 \\
    In-air FetchPickAndPlace & solved trajectory & SAC+HER & 100/50 & 3000 & 5/100 \\
    Meta-World PickPlace & solved trajectory & SAC+HER & 100/50 & 4500 & 6/100 \\
    \midrule
    PickAndPlace pacing conflict & solved trajectory & SAC+HER & 100/50 & 2400 & 6/100 \\
    Meta-World push $\times$ shelf-place & solved trajectory & SAC+HER & 200/100 & 6000 & 6/200 \\
    PickAndPlace within-group conflict & solved trajectory & SAC+HER & 100/50 & 2400 & 6/100 \\
    \bottomrule
  \end{tabular}
\end{table}

\begingroup
\raggedright
\paragraph{Task instances and cadence.} Environment IDs are
\path{PointMaze_Medium-v3}, \path{PointMaze_Large-v3},
\path{MiniGrid-MultiRoom-N4-S5-v0}, \path{MiniGrid-DoorKey-8x8-v0}, and
\path{FetchPickAndPlace-v4}. Meta-World uses \path{pick-place-v3}, or
\path{push-v3} plus \path{shelf-place-v3} with the \path{handnear:graspcur} modifier.
Episode-horizon/evaluation-interval pairs are $300/100$ for PointMaze, $200/50$ for MultiRoom,
$160/50$ for DoorKey, $50/150$ for Fetch, $150/100$ for Meta-World PickPlace, and $150/150$
for the Meta-World conflict.
\par
\endgroup

\paragraph{Scaffolds and observations.}\label{app:methods} Continuous tasks use five fractions
$\{0,.25,.5,.75,.9\}$; MiniGrid uses $\{0,.25,.5,.75,1\}$. PointMaze follows a geodesic,
MultiRoom a route, and DoorKey the key--door--goal stages. Manipulation restores states along a
solution trajectory, including a grasp for in-air starts. Level zero is always the original task.
Continuous learners receive low-dimensional goal-conditioned observations; MiniGrid uses a
partial-view convolutional encoder.

\paragraph{Evaluation and inference.} Held-out success is \emph{success-once}: any successful
step makes the episode successful. AUC integrates the held-out $\ell=0$ curve and is compared
only within a task. Main-paper natural-setting AUCs are seed medians; \cref{tab:auto} uses
condition-mean AUCs over a common block window; constructed and within-group tables use median
worst-group AUC; and the Meta-World conflict table uses seed means. Final columns report means
$\pm$ s.d. where shown. Paired conflict intervals bootstrap the median paired difference over
six seeds with $10{,}000$ resamples. Exact sign tests are descriptive: their six-seed resolution
is too coarse for any comparison to survive Holm correction over the full family. Bank builders,
seed IDs, available hashes, and archived exceptions are documented in the release materials.
Continuous evaluation uses two episodes for every held-out context; MiniGrid uses one greedy
episode on the fixed first $50$ of its $100$ held-out layouts at every checkpoint.

\clearpage
\subsection{Controller and learner implementation}

\paragraph{Controller semantics.}\label{app:hparams} One controller window is one training
iteration. Every context is checked once per window; a direction must remain eligible for two
consecutive windows before movement. Cell statistics persist across revisits, while unseen cells
cannot move before $n_{\min}$ observations. Sampling applies a softmax, mixes total mass
$\epsilon$ with the uniform distribution, caps each cell at $q_{\max}$, and water-fills excess
mass proportionally.

\begin{table}[H]
  \centering
  \caption{\textbf{Frozen controller constants}, shared across settings and conditions.}
  \label{tab:hparams}
  \small
  \begin{tabular}{@{}llc@{}}
    \toprule
    Component & Symbol & Value \\
    \midrule
    Sampling score weights (\cref{eq:score}) & $\lambda_h$ / $\lambda_p$ / $\lambda_{\mathrm{stale}}$ & $1.0$ / $0.2$ / $0.1$ \\
    Softmax temperature & $\tau$ & $0.5$ \\
    Exploration floor / per-cell cap & $\epsilon$ / $q_{\max}$ & $0.1$ / $0.5$ \\
    Transition thresholds (\cref{eq:transition}) & $\tau_{\mathrm{low}}$ / $\tau_{\mathrm{high}}$ & $0.15$ / $0.8$ \\
    Evidence gate & $n_{\min}$ & $15$ rollouts \\
    Confirmation / cooldown & --- & $2$ / $3$ windows \\
    Antistall limit & $N_{\mathrm{stale}}$ & $4$ \\
    Success EMA & $\beta$ & $0.2$ \\
    \bottomrule
  \end{tabular}
\end{table}

\begin{table}[H]
  \centering
  \caption{\textbf{Learner hyperparameters.} Values are shared across conditions; budgets are in
  \cref{tab:setup}. Setting-specific deviations are listed explicitly.}
  \label{tab:learner}
  \footnotesize
  \begin{tabular}{@{}>{\raggedright\arraybackslash}p{0.27\linewidth}>{\raggedright\arraybackslash}p{0.69\linewidth}@{}}
    \toprule
    Component & Value \\
    \midrule
    SAC architecture & MLP $(256,256)$, ReLU; double-Q critic; tanh-Gaussian actor \\
    SAC optimization & Adam, $3\times10^{-4}$; $\gamma=0.99$; target rate $0.005$; automatic entropy (initial $e^{-1.6}$, target $-|\mathcal A|$); batch $256$; warmup $1000$; running observation normalization \\
    HER / exploration & Future HER, $k=4$; random/action noise $0.1/0.1$ on PointMaze and $0.3/0.2$ otherwise \\
    SAC deviations & Replay $3\times10^5$ for PointMaze/Fetch, $4\times10^5$ for Meta-World PickPlace, $6\times10^5$ for its conflict; workers $6/16/8$; $16$ updates per iteration, or $32$ for Fetch \\
    PPO architecture & Conv $3{\to}16{\to}32$ ($2\times2$), MLP $(128,128)$, Tanh \\
    PPO optimization & Adam, $2.5\times10^{-4}$; $\gamma=0.99$; GAE $0.95$; clip $0.2$; entropy/value coefficients $0.01/0.5$; $4$ epochs; batch $1024$; $16$ episodes/iteration \\
    \bottomrule
  \end{tabular}
\end{table}

\paragraph{Compute and software.} Runs use CPU learners. The locked environment is Python
$3.12.13$, PyTorch $2.12.1$, NumPy $2.5.0$, Gymnasium $1.3.0$, Gymnasium-Robotics $1.4.2$,
MiniGrid $3.1.0$, Meta-World $3.0.0$, and MuJoCo $3.10.0$; \texttt{uv.lock} pins the full
environment. A $2400$-iteration Fetch conflict run takes
about $40$ minutes on a $20$-core DGX Spark and up to five hours on the cluster CPU nodes.

\subsection{Baselines and tuning provenance}
\label{app:protocols}
\begin{table}[H]
  \centering
  \caption{\textbf{Baseline reset-selection rules.} Learner, contexts, reward, rollout budget,
  and evaluation are fixed within each setting. $\dagger$ marks conflict-only conditions and
  $\ddagger$ oracle-label conditions.}
  \label{tab:baselines}
  \scriptsize
  \setlength{\tabcolsep}{4pt}%
  \begin{tabular}{@{}>{\raggedright\arraybackslash}p{0.22\linewidth}>{\raggedright\arraybackslash}p{0.73\linewidth}@{}}
    \toprule
    Condition & Reset-cell rule and constants \\
    \midrule
    Target start & Every rollout at $\ell=0$; no scaffold. \\
    Fixed level$^{\dagger}$ & Constant $\ell$; ``fixed mid'' is $\ell=2$. \\
    Random scaffold & $\ell\sim\mathrm{Unif}\{0,\ldots,L\}$ independently per rollout. \\
    Global annealing & $\ell(t)=\mathrm{round}[L\max(0,1-t/T)]$; $T=1500$ for PointMaze/MiniGrid, $2000$ for in-air Fetch and both Fetch conflicts, $2250$ for Meta-World PickPlace, and $3000$ for the Meta-World conflict. \\
    Synchronized RC & Continuous start progress $0.9\to0$ linearly over $2000/3000$ iterations. \\
    Global mixture$^{\dagger}$ & Aggregate-success EMA rate $.05$ grows $q(\ell=0)$ from $.1$ to $.7$; easiest-level mass $.2$; remainder uniform. \\
    RFCL-style local frontier & Our controller's transitions with uniform context sampling and antistall disabled. \\
    Group frontier$^{\dagger\ddagger}$ & One frontier and statistic per oracle group using our controller; context uniform within group. \\
    PLR cells$^{\dagger}$ & Rank-prioritized mean $|\mathrm{TD}|$ over all $(x,\ell)$ cells; $\beta=.5$, staleness mix $\rho=.1$, EMA $\alpha=.3$. \\
    ALP cells$^{\dagger}$ & $|s_{\mathrm{fast}}-s_{\mathrm{slow}}|+.1\log(1+\mathrm{stale})$ over all cells; EMA rates $.2/.05$, $\tau=.5$, and our controller's floor/cap. \\
    RFCL advance$^{\dagger}$ & Advance after $\geq3$ of the last $5$ successes; no retreat/antistall; sampling weight $(1-\hat s)+.1$. \\
    ACED-style advance$^{\dagger}$ & our controller's EMA, evidence gate, confirmation, cooldown, and uniform context sampling; advance at $\tau_{\mathrm{high}}$, with no retreat or antistall. \\
    Anneal + replay$^{\dagger\ddagger}$ & Global annealing with equal per-group SAC minibatch quotas; empty groups cede quota. \\
    \bottomrule
  \end{tabular}
\end{table}

\paragraph{RFCL protocols.} Official RFCL uses $100$ state-only demonstrations and a REDQ-style
ensemble at update-to-data ratio $10$ on PointMaze Large. Over three one-million-step seeds,
RFCL learns faster and reaches median final success $0.31$, versus $0.34$ for our controller with automatic scaffolds; target-start
SAC+HER remains at zero. Learners and seed sets differ, so this is an ecosystem comparison.
The conflict tables separately name the component-matched local-frontier and advance-only rules.

\paragraph{Tuning provenance.} Controller values were selected on development runs and frozen
before the reported tests. A full fixed-level screen identifies $\ell=2$ as the strongest
constant by worst-group AUC on both conflicts ($0.131$ constructed; $0.024$ Meta-World); no
constant exceeds $0.15$ Meta-World worst final. Across constructed-conflict annealing lengths
$\{800,1200,1600,2000,2400\}$, the largest median worst-group AUC is $0.0174$, so $T=2000$ is
not a favorable choice. Mixture coefficients were set a priori; group and cell teachers add no tuned
values. Final conflict comparisons use seeds $42$--$47$; screens use PointMaze seeds $40$--$42$,
constructed-conflict seeds $42$--$44$, or DoorKey seeds $50$--$52$. The constructed screen
therefore overlaps the final seed IDs and is sensitivity/provenance evidence, not an independent
test block.

\section{Automatic scaffold construction and validation}
\label{app:auto}
The extractor samples fixed progress quantiles from one successful trajectory per context.
PointMaze uses shortest-path plans; DoorKey uses a breadth-first solver and omits the terminal
goal-reaching step, leaving its easiest reset adjacent to the goal; and Fetch uses a four-phase
pick-and-place controller. PointMaze serializes point position and velocity, DoorKey the grid,
pose, and carried object, and Fetch the full MuJoCo state, actuator controls, and welded mocap
target needed to preserve an in-air grasp. Every extracted state is restored and settled before
training; all ladders pass finite-state, bounds, and task-specific grasp checks. DoorKey's fixed
quantiles recover the key--door--goal stages without event labels.

We summarize recovery of the hand-built scaffold's gain using
\begin{equation}
R_{\mathrm{retain}} \;=\;
\frac{\mathrm{AUC}_{\mathrm{auto}}-\mathrm{AUC}_{\mathrm{true}}}
     {\mathrm{AUC}_{\mathrm{manual}}-\mathrm{AUC}_{\mathrm{true}}},
\label{eq:rretain}
\end{equation}
Thus $R_{\mathrm{retain}}\geq1$ means that the automatic ladder matches or exceeds the manual
ladder's gain over target-start training.

\begin{table}[H]
  \centering
  \caption{\textbf{Trajectory-derived scaffolds retain hand-built benefits.} Entries are
  condition-mean held-out AUCs over each block's common window (three seeds for continuous tasks,
  six for DoorKey). Intervals bootstrap $R_{\mathrm{retain}}$; DoorKey's near-zero, bimodal
  denominator makes its interval unstable, so only the point estimate is shown.}
  \label{tab:auto}
  \small
  \begin{tabular}{lcccc}
    \toprule
    Setting & Target start & Ours (manual) & Ours (automatic) & $R_{\mathrm{retain}}$ [95\% CI] \\
    \midrule
    PointMaze Large & 0.012 & 0.225 & 0.269 & $1.21\;[1.05,1.44]$ \\
    MiniGrid DoorKey & 0.259 & 0.451 & 0.541 & $1.47$ \\
    In-air FetchPickAndPlace & 0.002 & 0.801 & 0.845 & $1.06\;[1.05,1.06]$ \\
    \bottomrule
  \end{tabular}
\end{table}

\section{Pacing-conflict theory and evidence}
\label{app:conflicts}
We prove the scheduling limitation below, then test whether the same issue appears in constructed and unmodified conflicts.

\subsection{Toy construction and proof}
\label{app:toy}
Consider contexts $a,b$ and levels $\ell=1$ (assisted), $\ell=0$ (target). Each context receives
$H=K+1$ episodes, $K>1$.
Let $n_1(x)$ and $n_0(x)$ denote the number of assisted and target-start episodes assigned to
context $x$. Assume context $a$ succeeds at evaluation iff $n_1(a)\ge 1$ and $n_0(a)\ge K$,
and context $b$ iff $n_1(b)\ge K$ and $n_0(b)\ge 1$. Since $n_0(x)=K+1-n_1(x)$, these reduce to
\[
    a \text{ succeeds iff } n_1(a)=1, \qquad b \text{ succeeds iff } n_1(b)=K .
\]
Context $a$ needs assistance removed quickly; context $b$ needs it retained.

\begin{mainprop}[synchronized global pacing can be insufficient; restated]
For every $K>1$, there exists a two-context, two-level scaffold curriculum with per-context budget $K+1$
such that a context-local schedule solves both contexts, while every synchronized global schedule fails at
least one context.
\end{mainprop}

\begin{proof}
A synchronized global schedule gives both contexts the same assisted count $n_1$; context $a$
requires $n_1=1$ and context $b$ requires $n_1=K>1$, so every synchronized schedule fails at
least one context. A context-local schedule assigns $n_1(a)=1$ and $n_1(b)=K$, satisfying both
under the same per-context budget.
\end{proof}

\paragraph{Scope.} The proof applies to synchronized schedules. Global mixtures remain possible, and the result makes no claim about the optimality of our controller.

\subsection{Constructed and within-group pacing conflicts}
\label{app:extra}
\cref{tab:pnp_c5} uses each experiment's common six-seed window. The constructed panel uses the
archived replication block; the main-paper headline retains the earlier $0.354/0.123$ point
estimates, whose raw per-seed block is not part of the local analysis cache.

\begin{table}[H]
  \centering
  \caption{\textbf{Conflict comparisons across six seeds.} AUC is the median worst group/band;
  final is mean $\pm$ s.d.; $\Delta$ is the median paired ours-minus-condition difference with
  a bootstrap 95\% CI. Panel (a) uses the constructed replication block. $\ddagger$ uses oracle labels.}
  \label{tab:pnp_c5}
  \scriptsize
  \setlength{\tabcolsep}{4pt}%
  \begin{tabular}{lccc}
    \toprule
    Condition & Worst AUC & Worst final & $\Delta$ [95\% CI] \\
    \midrule
    \multicolumn{4}{@{}l}{\emph{(a) Constructed conflict}} \\
    \PCRCPerContextSuccess{} & \textbf{0.378} & $0.97 \pm 0.04$ & (ref.) \\
    Group frontier$^{\ddagger}$ & 0.229 & $0.95 \pm 0.02$ & $+0.18\;[0.04,0.29]$ \\
    RFCL advance & 0.297 & $\mathbf{0.98 \pm 0.03}$ & $+0.13\;[-0.09,0.30]$ \\
    ACED-style advance & 0.083 & $0.60 \pm 0.24$ & $+0.30\;[0.23,0.40]$ \\
    PLR cells & 0.239 & $0.63 \pm 0.37$ & $+0.15\;[0.05,0.35]$ \\
    ALP cells & 0.123 & $0.81 \pm 0.36$ & $+0.25\;[0.13,0.40]$ \\
    \midrule
    \multicolumn{4}{@{}l}{\emph{(b) Within-group conflict under one label}} \\
    \PCRCPerContextSuccess{} & \textbf{0.701} & $0.91 \pm 0.10$ & (ref.) \\
    \PCRCPerGroupSuccess{}$^{\ddagger}$ & 0.000 & $0.22 \pm 0.35$ & $+0.69\;[0.47,0.74]$ \\
    RFCL advance & 0.396 & $0.93 \pm 0.08$ & $+0.30\;[0.17,0.48]$ \\
    ACED-style advance & 0.553 & $0.95 \pm 0.07$ & $+0.16\;[0.11,0.27]$ \\
    PLR cells & 0.615 & $0.92 \pm 0.14$ & $+0.08\;[0.05,0.18]$ \\
    ALP cells & 0.551 & $0.85 \pm 0.20$ & $+0.14\;[0.06,0.25]$ \\
    Global annealing & 0.527 & $\mathbf{0.96 \pm 0.06}$ & $+0.20\;[0.09,0.27]$ \\
    Global mixture & 0.451 & $0.83 \pm 0.18$ & $+0.29\;[0.17,0.39]$ \\
    \bottomrule
  \end{tabular}
\end{table}

In the primary block, the asynchronous mixture reaches only $0.022$ median worst-group AUC.
With automatic ladders, our controller also improves sample efficiency over annealing ($0.73$ versus
$0.55$); both eventually solve both groups. In panel (b), the low-goal band moves the pooled
frontier prematurely, leaving zero high-band success on four of six seeds.

\begin{figure}[H]
  \centering
  \includegraphics[width=\linewidth]{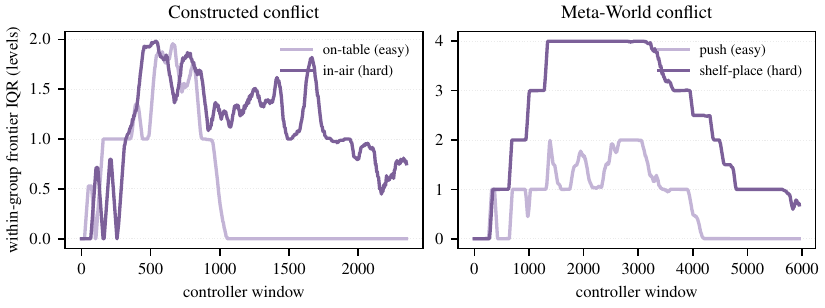}
  \caption{\textbf{Within-group frontier dispersion} (median across six seeds; smoothed).
  Light and dark curves denote easy and hard groups. Our controller's frontiers disperse most within hard
  groups.}
  \label{fig:c5variance}
\end{figure}

\paragraph{Within-group frontier dispersion.} Frontiers in easy groups mostly converge. Hard groups show much more spread: one to two levels in the constructed conflict and as many as four on shelf-place (\cref{fig:c5variance}). A pooled frontier cannot represent this variation.

\subsection{Meta-World conflict and robustness}
\label{app:conflict-search}
Push can learn despite the assisted starts; shelf-place needs rollouts that begin after grasping. We combine
$100$ training and $50$ held-out contexts per task for $6000$ iterations over six seeds. Our controller
advances push early while retaining assistance for shelf-place; \cref{app:protocols} defines all
comparison rules.

\begin{table}[H]
  \centering
  \caption{\textbf{Meta-World push--shelf-place conflict.} Six-seed means after $6000$
  iterations with five levels; worst AUC minimizes across tasks. $\ddagger$ uses oracle labels.}
  \label{tab:mw-conflict}
  \footnotesize
  \begin{tabular}{lccc}
    \toprule
    Condition & Worst AUC & Push final & Shelf final \\
    \midrule
    \PCRCPerContextSuccess{} & 0.141 & 0.743 & 0.617 \\
    Group frontier$^{\ddagger}$ & \textbf{0.256} & 0.840 & \textbf{0.757} \\
    RFCL-style local frontier & 0.164 & 0.743 & 0.373 \\
    RFCL advance & 0.042 & 0.763 & 0.167 \\
    ACED-style advance & 0.049 & 0.777 & 0.313 \\
    PLR cells & 0.118 & 0.683 & 0.197 \\
    ALP cells & 0.140 & 0.823 & 0.477 \\
    \midrule
    Global annealing & 0.003 & 0.843 & 0.000 \\
    Anneal + balanced replay$^{\ddagger}$ & 0.006 & 0.803 & 0.000 \\
    Global mixture & 0.125 & 0.757 & 0.383 \\
    Random scaffold & 0.117 & 0.803 & 0.477 \\
    Fixed mid & 0.022 & 0.753 & 0.053 \\
    Target start & 0.000 & \textbf{0.880} & 0.000 \\
    \bottomrule
  \end{tabular}
\end{table}

\paragraph{Horizon and ladder resolution.} More training does not rescue annealing with five scaffold levels, although the mixture catches up. With nine levels, all three methods solve both tasks (\cref{tab:mw-robust}).

\begin{table}[H]
  \centering
  \caption{\textbf{Horizon and ladder-resolution robustness.} Cells report worst-group AUC /
  shelf-place final. The two-seed, nine-level row is exploratory.}
  \label{tab:mw-robust}
  \small
  \begin{tabular}{lccc}
    \toprule
    Protocol (ladder, budget, seeds) & \PCRCPerContextSuccess{} & Global annealing & Global mixture \\
    \midrule
    5 levels, 6000 it, 6 seeds & 0.141 / 0.617 & 0.003 / 0.000 & 0.125 / 0.383 \\
    5 levels, 9000 it, 3 seeds & 0.337 / 0.680 & 0.008 / 0.000 & 0.347 / 0.853 \\
    9 levels, 9000 it, 2 seeds & 0.367 / 0.840 & 0.353 / 0.890 & 0.374 / 0.810 \\
    \bottomrule
  \end{tabular}
\end{table}

\section{Controller sensitivity and boundary conditions}
\label{app:additional}

\subsection{Controller sensitivity}
The frozen $N_{\mathrm{stale}}=4$ predates a retrospective three-seed sensitivity check and was
not selected by it. Median worst-group AUC for $N_{\mathrm{stale}}=0,2,4,8$ was respectively
$0.15,0.49,0.20,0.39$. Each enabled setting reached $1.00$ on-table final and $0.92$--$1.00$
hard final, versus $0.96/0.84$ with antistall disabled. The enabled values are nonmonotonic and
do not establish that four is optimal.

\paragraph{Alternative transition rules.} The Wilson-gated variant uses Wilson bounds over discounted success
counts and the Beta-posterior variant a discounted Beta posterior. Across six automatic-conflict seeds, these rules,
our controller, and the RFCL-style local frontier have similar worst-group AUC; evidence gating roughly
halves direction reversals. Old evidence must decay: an undiscounted posterior
falls to $0.58$ worst-group AUC. Across these variants, the local frontier is what holds up. The specific thresholding rule matters less.

\subsection{When scaffolds and local pacing help less}
\label{sec:res-strong}
Three exploratory studies bound the claims. On long-horizon DoorKey-16$\times$16 (three seeds),
increasing the automatic ladder from five to nine levels raises median final success from $0.72$
to $0.82$ without reducing AUC; thirteen levels reach $0.84$ final but reduce AUC from $0.62$ to
$0.49$. On two three-seed solution-length splits, our controller has about twice annealing's worst-group
AUC, while the asynchronous mixture matches or exceeds our controller, consistent with strong cross-level
transfer.

A three-seed comparison with our exploratory Torch contrastive critic
\citep{eysenbach2022contrastive} uses in-batch InfoNCE negatives, a
$64$-dimensional representation, a $0.1$ log-sum-exp penalty, HER-future positives, and the SAC
actor. With this learner, target-start/ours final success is $0.48/0.41$ on PointMaze Large and
$0.20/0.24$ on in-air Fetch; our controller with the standard SAC+HER learner reaches $1.00$ on Fetch.
A directed learner can sometimes overcome the state-access problem on its own. Strong transfer across scaffold levels can also reduce the need for per-context pacing.

\end{document}

%% file: generated/names.tex
\newcommand{\PCRCTitle}{Per-Context Reverse Curricula for Sparse-Reward Reinforcement Learning}
\newcommand{\PCRCPerContextSuccess}{Per-context success}

\newcommand{\PCRCPerGroupSuccess}{Per-group success}

%% file: tables/sixtask_table.tex
\setlength{\tabcolsep}{4pt}
\begin{tabular}{llcccccccc}
\toprule
Task & Learner & Target & Global & Random & Per-context success & $\Delta$ O$-$T & Fin.\ T & Fin.\ O & Solv.\ O/T \\
 & & AUC & AUC & AUC & AUC & AUC &  &  &  \\
\midrule
PointMaze Medium & SAC+HER & 0.254 & 0.559 & 0.407 & 0.559 & +0.314 & 0.62 & 0.63 & 5/5 \\
PointMaze Large & SAC+HER & 0.019 & 0.230 & 0.137 & 0.245 & +0.226 & 0.19 & 0.30 & n/a \\
MiniGrid MultiRoom & PPO & 0.000 & 0.614 & 0.440 & 0.633 & +0.633 & 0.00 & 1.00 & 6/0 \\
MiniGrid DoorKey & PPO & 0.000 & 0.000 & 0.363 & 0.300 & +0.268 & 0.46 & 0.95 & 6/3 \\
In-air FetchPickAndPlace & SAC+HER & 0.002 & 0.791 & 0.789 & 0.840 & +0.838 & 0.00 & 1.00 & 5/0 \\
Meta-World PickPlace & SAC+HER & 0.028 & 0.298 & 0.167 & 0.240 & +0.172 & 0.02 & 0.73 & 6/0 \\
\bottomrule
\end{tabular}